# Clustering of Spell Variations for Proper Nouns Transliterated from the other languages

Prathamesh Pawar

*Khoury College of Computer Sciences,*

*Northeastern University*

Boston, United States

pawar.prath@northeastern.edu

**ABSTRACT**

**One of the prominent problems with processing and operating on text data is the non uniformity of it. Due to the change in the dialects and languages, the caliber of translation is low. This creates a unique problem while using NLP in text data; which is the spell variation arising from the inconsistent translations and transliterations. This problem can also be further aggravated by the human error arising from the various ways to write a Proper Noun from an Indian language into its English equivalent. Translating proper nouns originating from Indian languages can be complicated as some proper nouns are also used as common nouns which might be taken literally. Applications of NLP that require addresses, names and other proper nouns face this problem frequently. We propose a method to cluster these spell variations for proper nouns using ML techniques and mathematical similarity equations. We aimed to use Affinity Propagation to determine relative similarity between the tokens. The results are augmented by filtering the token-variation pair by a similarity threshold. We were able to reduce the spell variations by a considerable amount. This application can significantly reduce the amount of human annotation efforts needed for data cleansing and formatting.**

## I. INTRODUCTION

Proper Nouns translated from languages belonging to the Indian subcontinent have been quite difficult to spell check due to the nature of these words. This poses a challenge while working with any system that deals with names of geographical locations such localities, villages or districts. A significant number of databases in India rely on manually transliterated English versions of these words. Such transliterations are prone to errors or variations due to the subjectivity of the task. The similarity between two words is more than the similarity between a word and its misspelled version. The existence of common suffixes and prefixes further aggravates the problem. We propose a method to cluster the misspelled versions of words using Affinity propagation algorithm and jaro-winkler similarity. Affinity propagation is a graph based clustering algorithm. It does not require a predetermined number of clusters to run the algorithm. It uses a series of similarities and matrices that we use in conjunction with the jaro-winkler distance algorithm as a thresholding system to improve the results. Affinity propagation

## II. RELATED WORKS

Amorim et al. [1] used Clustering Algorithms for spell checking. They experimented with 36,133 words with 6,136 target words. They implemented it using PAM (Partition Around Medoids) and Anomalous Pattern Initialization. The PAM method proposed by Van der Laan et al. [2] is used to create clusters at random and assign a certain medoid for each of them. Then an optimum solution is derived by minimizing the net distance between the clusters and their respective medoids. This method was then improved by using Anomalous Pattern Initialization to initialize the medoids with specific values. This increases the accuracy as the medoids don't depend on random initializations. The words were also divided into 26 initial clusters based on the first alphabetical order in order to reduce the processing load on the machine. This method achieved an impressive accuracy of 88.42%. However, there is no equivalent implementation in Indian transliterated language words.

## III. PROPOSED METHOD

I have proposed a method based on Affinity Propagation and jaro-winkler similarity.

### A. Affinity Propagation

Affinity Propagation is a clustering algorithm proposed by Frey et al. [3] that uses a message based system which informs the tokens about the relative attractiveness of each other. The goal of the algorithm is to find an exemplar for a particular token which is the best representation of that token. To calculate the exemplar we first generate a nxn similarity matrix where n is the number of tokens in the dataset. The matrix is created by calculating the similarity between every token with every other token. Levenshtein distance is used to calculate the distance [4]. To make sure that the tokens do not assign themselves as their exemplar we make all the diagonal values as $min(sim(i, j))$

$$S(i, j) = -||x_i - x_j||^2$$

Calculating similarity

We further calculate the Responsibility Matrix which is basically determining how suitable is a token k as an exemplar for token i with respect to the closest alternative token k'.

$$r(i, k) \leftarrow s(i, k) - \max_{k' \neq k} \{a(i, k') + s(i, k')\}$$

Calculating responsibility

The Responsibility Matrix is initialized as a nxn matrix with zeros. Each position in the matrix is populated by the equation. Now that we have determined if the token k would be an appropriate exemplar for i, we now determine if the token i would be a suitable cluster member for exemplar k with respect to other tokens. This is to be calculated using the Availability Matrix.

$$a(i,k) \leftarrow \min\left(0, r(k,k) + \sum_{i' \notin \{i,k\}} \max(0, r(i',k))\right) \text{ for } i \neq k$$

Calculating Availability

The Availability is dependent on the positive responsibilities and the self availability of each token.

$$a(k,k) \leftarrow \sum_{i' \neq k} \max(0, r(i',k)).$$

Self Availability

Criterion Matrix is the final stage of the Algorithm. It is the sum of the responsibility matrix and availability matrix.

$$c(i,k) \leftarrow r(i,k) + a(i,k).$$

Criterion Matrix

The exemplar is determined by the highest value in the criterion matrix.

### B. Jaro-Winkler Distance

In order to increase the number of tokens successfully, there are no checks imposed on the tokens assigned to a cluster. This hampers the accuracy of clusters in order to accommodate a wide range of tokens. Therefore to improve the accuracy, we introduce a threshold to filter the tokens that are not similar to the exemplar of that cluster. In order to introduce this threshold in the clusters we use Jaro-Winkler Similarity.

$$sim_j = \begin{cases} 0 & \text{if } m = 0 \\ \frac{1}{3}\left(\frac{m}{|s_1|} + \frac{m}{|s_2|} + \frac{m-t}{m}\right) & \text{otherwise} \end{cases}$$

Jaro similarity

m = no. of common letters    t = no. of transpositions required

$$sim_w = sim_j + \ell p(1 - sim_j)$$

Jaro-Winkler Similarity

simj = Jaro Similarity    lp = length of common prefix

Jaro Winkler Similarity [5] is a modified version of Jaro similarity where the similarity is calculated by introducing common prefix length weight to the similarity. We use this similarity as a threshold to maintain the sanctity of the cluster.

## IV. Data and Preprocessing

### A. Data Extraction

The goal was to make this process as practical as possible. In order to achieve that a dataset was obtained from government records of the deed records from Delhi real estate property addresses and extract localities from them. Since these records are manually entered we found a large amount of spelling variations of the extracted localities. For Eg. *Mehrauli* was represented as *Mehroli, Mehrali, Mehrouli* etc We annotate the dataset to form the perfect clusters manually. The performance of the algorithm will be tested against this baseline. We also replicate the experiment with 2 more datasets. We use Mumbai Apartment names which have been obtained from public deed records of Mumbai. We use these datasets as they are a perfect representation of the problem that we face while translating the nouns. A literal translation cannot be done as they might distort the meaning of the word by taking its literal meaning instead of the word itself. While these datasets have been transliterated from Devanagari script which may not be the case for other Indian languages, it is representative of the problem that arises from transliterating [6] from them.

An important aspect of these datasets is that they are generated through public records which are manually filled out and thus have a significant number of spelling errors in their original text which in turn results in spell errors in transliterated text.

| Dataset | Size |
|---|---|
| Spell (F1) | 406 |
| Mumbai Apartments (D1) | 16295 |
| Delhi Localities (D2) | 13120 |

### B. Preprocessing

In order to make sure that the clustering algorithm can function efficiently, we clean the phrases and tokens.. We take out the common parts of a phrase that do not contribute to the uniqueness. For Eg. In *Anand Nagar,* the word *Nagar* is common with many other tokens and might make the algorithm askew. All the abbreviations and numbers are also removed. It is assumed that the probability of the spelling variations occurring in the first letter. So, to increase the efficiency of the algorithm, we divide all the tokens into 26 (A-Z) groups according to the first letter of the tokens.

## IV. Procedure

1. We first create the similarity matrix by calculating the similarity matrix using the levenshtein distance.
2. We implement affinity propagation on the matrix to generate clusters containing tokens with their respective spelling variations.
3. Damping Factor is a very important parameter that prevents the availability matrix and responsibility matrix from overshooting the results which may lead into oscillations instead of a straight convergence. For this particular application we keep it to 0.65.
4. After implementing Affinity Propagation, a dictionary is generated with clusters and their respective exemplars.
5. In order to filter the false positives from the clusters, the Jaro-Winkler Distance Threshold is implemented.

6. Every element of the cluster is compared with the exemplar of that cluster to calculate the Jaro-Winkler Distance.
7. A threshold of 95% similarity is applied to keep the ones that are very close to the exemplar to remove all the False positives that we can.
8. While the majority of False positives are eliminated, a significant portion of the tokens are still not clustered. Inorder to utilize these tokens, the algorithm is run again on the remaining clusters.
9. The cluster dictionaries are combined to form a final cluster dictionary.

## V. Evaluation Metrics

Evaluating the performance of a clustering algorithm is fairly convoluted. The number of clusters created can be varied based on the given data and parameter values. In order to evaluate, we first create a set of true clusters by annotation. These clusters shall be used to create token-variation pairs which would essentially be pairs of each cluster element and the corresponding exemplar.

In order to evaluate the performance of a clustering algorithm we use a mathematical method known as Adjusted Rand Index [8]. The ARI is used because it gives us the most realistic evaluation for the resultant clusters. ARI takes into account the Expected value of the clustering algorithm i.e. the likelihood of token-variations pairs being correct on random.

In order to calculate ARI we first calculate the Rand Index (RI) [7] which is calculated as

$$RI = \frac{TP + TN}{TP + FP + FN + TN}$$

- **TP** is the number of token-variation pairs which are the same in predicted and true pairs (True Positives).
- **TN** is the number of token-variation pairs that are neither in predicted pairs nor in true pairs.
- **FP** is the number of token-variation pairs which are in predicted tokens but not in true true pairs.
- **FN** is the number of token-variation pairs which are not in predicted tokens but are in true pairs.

After calculating the Rand Index, ARI is calculated using the equation.

$$ARI = \frac{RI - \text{Expected RI}}{\text{Max}(RI) - \text{Expected RI}}$$

| Dataset | Spellings (F1) | Mumbai Apartments (D1) | Delhi Locality (D2) |
|---|---|---|---|
| TP | 332 | 6296 | 6596 |
| FP | 2 | 246 | 285 |
| TN | 36 | 8506 | 5662 |
| FN | 5 | 1247 | 577 |
| RI | 0.999 | 0.998 | 0.995 |
| ARI | 0.946 | 0.934 | 0.928 |

Results

## VI. Conclusion

The subjectivity in translating proper nouns from various languages in English poses a big challenge in processing data and performing NLP techniques on them. Based on the results of this application, clustering the tokens using Affinity Propagation and Jaro-Winkler similarity has proven to be an effective way to rectify the spell errors and spell variations. This could be used as a viable solution for many NLP problems as a way to cleanse and simplify the data. One of the shortcomings of this approach is the scalability. However, this can be mitigated by implementing parallel processing to measure similarity between tokens simultaneously and developing certain criteria to divide the tokens into smaller sets.

## VII. References

[1] Amorim, Renato & Zampieri, Marcos. (2013). Effective Spell Checking Methods Using Clustering Algorithms. International Conference Recent Advances in Natural Language Processing, RANLP.
[2] Mark Van der Laan, Katherine Pollard & Jennifer Bryan (2003) A new partitioning around medoids algorithm, Journal of Statistical Computation and Simulation, 73:8, 575-584, DOI: 10.1080/0094965031000136012
[3] Brendan J. Frey; Delbert Dueck (2007). "Clustering by passing messages between data points". Science. 315 (5814): 972–976.
[4] Levenshtein, Vladimir I. (February 1966). "Binary codes capable of correcting deletions, insertions, and reversals". Soviet Physics Doklady. 10 (8): 707–710
[5] Wang Y. "Efficient Approximate Entity Matching Using Jaro-Winkler Distance". International Conference on Web Information Systems Engineering
[6] Bhat I.A., Mujadia V., Tammewar A., Bhat R.A., Shrivastava M. IIIT-H system submission for FIRE2014 shared task on transliterated search Proceedings of the forum for information retrieval evaluation, FIRE '14, 978-1-4503-3755-7, ACM, New York, NY, USA (2015), pp. 48-53
[7] W. M. Rand (1971). "Objective criteria for the evaluation of clustering methods". Journal of the American Statistical Association. American Statistical Association. 66 (336): 846–850
[8] Hubert, L. and Arabie, P. (1985) Comparing partitions. Journal of Classification, 193–218